\documentclass[11pt]{article}
\usepackage[final]{acl}
\usepackage{times}
\usepackage{latexsym}
\usepackage[T1]{fontenc}
\usepackage[utf8]{inputenc}
\usepackage{microtype}
\usepackage{inconsolata}
\usepackage{graphicx}
\usepackage{lineno}
\definecolor{darkblue}{rgb}{0, 0, 0.5}
\hypersetup{colorlinks=true, citecolor=darkblue, linkcolor=darkblue, urlcolor=darkblue}
\usepackage[most]{tcolorbox}

\definecolor{CB_lightCyan}{HTML}{99DDFF}
\definecolor{CB_darkCyan}{HTML}{66CCFF}

\definecolor{CB_pear}{HTML}{BBCC33}
\definecolor{CB_pink}{HTML}{E4CEE1}
\usepackage{xspace}
\usepackage{subcaption}
\usepackage{multirow}
\usepackage{booktabs}
\usepackage{tabularx}  
\usepackage{url}
\usepackage{tcolorbox}
\tcbuselibrary{skins}
\usepackage{makecell}
\usepackage{enumerate}     
\usepackage{amsmath}       
\usepackage{enumitem}
\usepackage[normalem]{ulem}
\usepackage{graphicx}
\newcommand{\foxglove}{\raisebox{-0.3ex}{\includegraphics[height=1.2em]{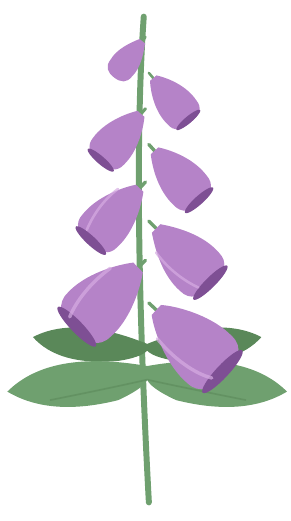}}}
\usepackage{booktabs}      
\usepackage{makecell}      
\usepackage{pifont}        
\usepackage{pifont}    
\usepackage{xcolor}    
\definecolor{cbteal}{RGB}{0, 77, 64}
\definecolor{cbpink}{RGB}{216, 27, 96}
\newcommand{\cmark}{\textcolor{cbteal}{\ding{51}}}
\newcommand{\xmark}{\textcolor{cbpink}{\ding{53}}}

\title{FOXGLOVE: Understanding Goal-Oriented and Anchored Writing \\ Feedback from Experts and LLMs on Argumentative Essays}
\newcommand{\dataset}{\foxglove\xspace\textsc{foxglove}\xspace} 
\author{
  Yijun Liu\textsuperscript{1}, Yifan Song\textsuperscript{1}, John Gallagher\textsuperscript{2}, 
  Sarah Sterman\textsuperscript{1}, Tal August\textsuperscript{1} \\
  \textsuperscript{1}Siebel School of Computing and Data Science \quad
  \textsuperscript{2}Department of English \\
  University of Illinois Urbana-Champaign \\
  Urbana, IL, USA \\
  \texttt{\{yijun6, yifan33, johng, ssterman, taugust\}@illinois.edu}
}

\tcbset{taggingPrompt/.style={
    enhanced,
    size=fbox,
    boxrule=2pt,
    arc=2mm,
    auto outer arc,
    left=1pt,
    right=1pt,
    top=1pt,
    bottom=1pt,
    fontupper=\ttfamily, 
    colback=CB_pink!15,
    colframe=CB_pink!30,
    coltitle=CB_pink!25!black, 
}}

\begin{document}
\maketitle
\begin{abstract}
While large language models (LLMs) are increasingly used to generate writing feedback, there remains no systematic comparison of LLM and expert feedback on the dimensions that writing research identifies as central to revision: goal-orientation, anchoring to specific sentences, and prioritization. We introduce FOXGLOVE, a dataset of 696 feedback comments written by trained writing instructors on 69 twelfth-grade argumentative essays, paired with 1,644 comments generated from four widely used LLMs under a shared protocol, totaling 2,340 comments. We find that instructors and LLMs distribute feedback similarly across argumentative goals and essay positions, yet diverge on the specific sentences on which to provide feedback. Additionally, we find that models tend to write more complex feedback and use fewer questions than instructors. We provide expert quality ratings on a subset of both instructor and LLM comments. In this preliminary expert quality rating study, we find that LLM feedback also receives higher ratings on most dimensions of quality, but much of this advantage appears to be attributable to lengthier comments. FOXGLOVE enables systematic comparison of where human and LLM feedback align, diverge, and differ.
\end{abstract}

\section{Introduction}

Writing feedback shapes how students develop as writers \cite{mah_christopher_sentence-corrections_nodate}, and that feedback is increasingly written by machines. Large language models (LLMs) can now produce feedback that engages with multi-dimensional analytic assessments \cite{wang_llms_2025}, including student writings' richness, visualization, interactivity, and personalization \cite{guerraoui_teach_2023}. Such feedback has been integrated into classroom workflows alongside teachers and teaching assistants \cite{lu_ai-mediated_2026, pahi_enhancing_2024, zhang_beyond_2026}, including deployed tools now used by hundreds of thousands of U.S. students \cite{khan_academy_annual_2025}. However, for writers and instructors to rely on generated feedback, there need to be effective ways of determining how LLM feedback differs from the feedback instructors traditionally give. Decades of writing research converge on what effective feedback should do:  it engages the rhetorical goals a writer is pursuing \cite{flower_cognitive_1981} and points to the specific text where revision should occur \cite{churchill_anchored_2000, weng_asynchronous_2004, zyto_successful_2012} while prioritizing the most important revisions \cite{sommers_responding_1982}. 

Whether human and model feedback align with one another on these criteria is unknown. This motivates our central question: \textbf{when humans and LLMs respond to the same essays under the same instructions, do they agree on where revision is needed, which argumentative goal should be addressed, and what feedback is the most urgent?}

\begin{figure}
    \centering
    \includegraphics[width=1\linewidth]{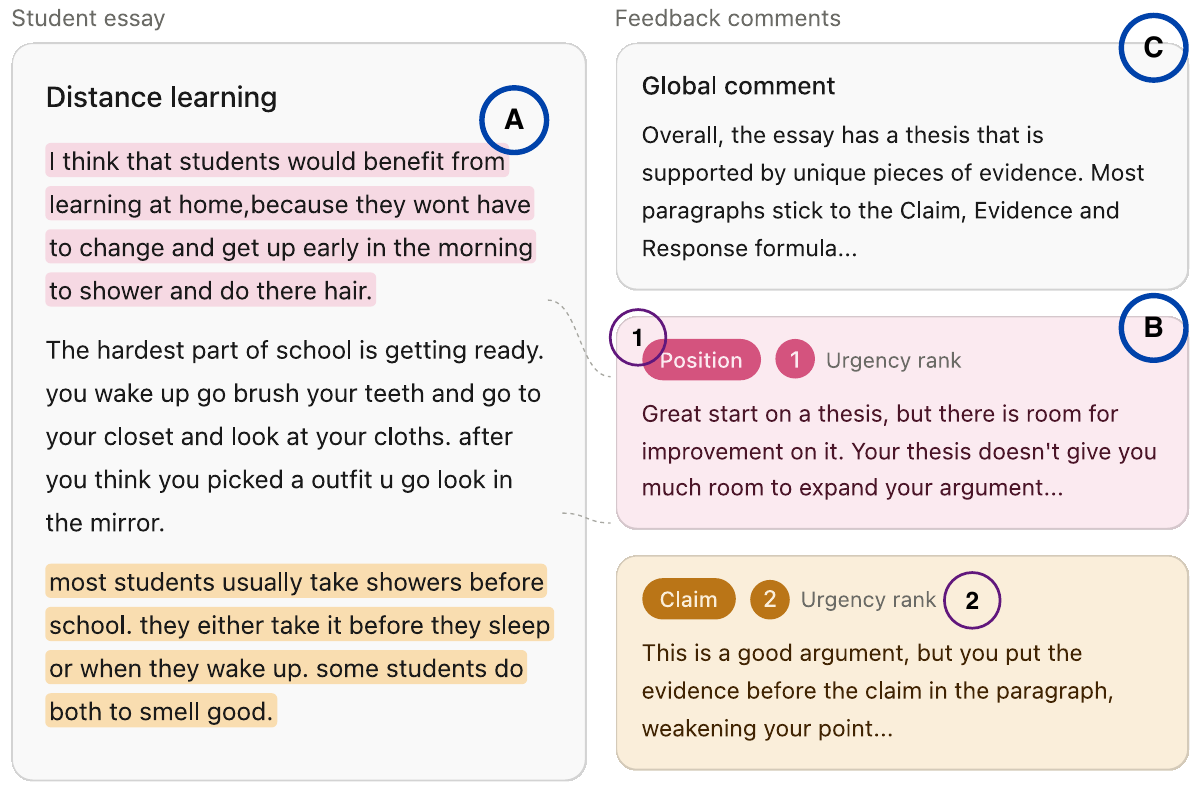}
    \caption{An illustrative essay showing the structure of \dataset that includes highlighted spans (A) linked to anchored feedback cards (B, goal label-1, urgency rank-2, and feedback comment) and one global comment (C) summarizing all the feedback comments.}
    \label{fig:intro}
\end{figure}

\begin{table*}[t]
\centering
\small
\setlength{\tabcolsep}{4pt}
\renewcommand{\arraystretch}{1.15}
\begin{tabular}{lcccccc}
\toprule
Dataset & \makecell{Anchored} & \makecell{Feedback\\Labels} & \makecell{Human\\Feedback} & \makecell{LLM\\Feedback} & \makecell{Urgency\\Rank} & \makecell{Human-Rated\\Quality} \\
\midrule
Expos\'ia \cite{zyska_exposi_2026}              & \cmark & \cmark  & Peer \& Instructor     & \cmark           & \xmark & \xmark \\

\citet{coyne_annotating_2025}   & \cmark & \xmark           & English Teacher                 & \cmark           & \xmark & \cmark \\

LEAF++ \cite{misgna_leaf_2026}                 & \xmark & \cmark  & Online Forum           & \cmark           & \xmark & \xmark \\

LEAF \cite{behzad_leaf_2024}                   & \xmark & \xmark         & Online Forum           & \cmark           & \xmark & \xmark \\

SEFL \citep{zhang_sefl_2026}            & \xmark & \xmark           & \xmark                 & \cmark           & \xmark & \cmark \\

\citet{mah_christopher_sentence-corrections_nodate}     & \cmark & \xmark           & English Teacher                 & \cmark           & \xmark & \xmark \\

\citet{pilan_dataset_2020}        & \cmark & \xmark         & Tutor                 & \xmark           & \xmark & \xmark \\
ArgRewrite v2 \cite{kashefi_argrewrite_2022}          & \xmark & \cmark         & Trained Expert                 & \xmark           & \xmark & \xmark \\

\citet{rashkin_help_2025} & \xmark & \xmark           & \xmark                 & \cmark           & \xmark & \cmark \\

\midrule
\textbf{\dataset} & \cmark & \cmark & Instructor & \cmark & \cmark & \cmark \\
\bottomrule
\end{tabular}
\caption{Comparison of multiple datasets on writing feedback.}

\label{tab:dataset_comparison}
\end{table*}

While past work has qualitatively described these differences \cite{mah_christopher_sentence-corrections_nodate}, answering this question at scale requires data that is presently unavailable (Table~\ref{tab:dataset_comparison}): there are no datasets that capture trained instructor and LLM generated feedback along the same protocol, anchored to specific text spans, labeled with the goals the feedback addresses, and ranked by urgency. 

We address this gap by introducing \dataset,\footnote{\dataset: \textbf{FO}rmative feedback from e\textbf{X}perts, with \textbf{G}oal-oriented, \textbf{L}ocalized Annotati\textbf{O}ns and e\textbf{V}aluations of student \textbf{E}ssays; \url{https://github.com/yijunliu23/foxglove_data_release}} a dataset we collect with anchored, goal-labeled feedback comments (Figure \ref{fig:intro}) from trained human instructors (696 comments) and four models (1{,}644 comments from GPT-5.2, Claude Sonnet 4.5, Llama 3.3 70B, and Qwen3-Next 80B) produced under a shared protocol. With the dataset, we systematically compare instructor and LLM feedback. Our paper makes the following contributions:

\begin{itemize}
    \item \textbf{A parallel feedback dataset} of 2{,}340 comments on 69 12th grade argumentative essays \cite{crossley_large-scale_2024}, produced by 14 expert writing instructors and a shared instructional protocol for eliciting model feedback in the same form. We document and release procedures for replicating instructor annotation and extending to additional models.
    
    \item \textbf{Analysis of instructor and LLM comments} showing where instructors and models place feedback across goals and essay positions, how they rank urgency, and how their language differs. We find that instructors and LLMs distribute feedback similarly across argumentative goals and essay positions, yet models generally generate more feedback and longer comments, and agree only modestly with instructors on the exact sentences to flag and rarely on how urgent feedback is. A prompting ablation shows that stylistic differences such as question use and personal address are largely steerable, while shortened comments become denser rather than simpler.

    \item \textbf{Independent expert quality ratings} on 105 feedback comments across 7 dimensions from two trained writing instructors, constituting a total of 1{,}430 quality ratings on both instructor and model feedback. In this preliminary study, LLMs generally receive higher scores in all dimensions except for tone; however, much of this advantage can be accounted for by LLMs' longer comments. 
\end{itemize}
\section{The Shifting Role of LLMs in Writing Feedback}

Prior studies suggest that LLM feedback can improve revision quality and motivation \citep{llm_classroom_meyer, zhang_beyond_2026}. Hybrid human--AI workflows can further improve feedback quality, with AI-mediated feedback sometimes outperforming human-only feedback in revision outcomes and clarity \citep{lu_ai-mediated_2026, pahi_enhancing_2024}. Carefully prompted open-weight models can also produce rubric-aligned feedback approaching expert quality in some settings \citep{allen_can_2026}.

However, LLM feedback also has clear limitations. It may overemphasize surface-level corrections, use imperative and non-dialogic language, miss the most important issue in a draft, or produce critiques that lack depth and cluster around similar text segments \citep{mah_christopher_sentence-corrections_nodate, rashkin_help_2025, liang_can_2023}. Comparative work generally finds that LLM feedback is more encouraging, structured, and consistently actionable, while human feedback is often more selective, specific, empathetic, and attentive to higher-priority issues \citep{steiss_comparing_2024, behzad_assessing_2024, liu_comparative_2025}. These results suggest that LLM feedback is useful at scale, but human judgment remains important for writing-intensive contexts requiring personalization and prioritization \citep{liu_comparative_2025}.

Existing writing-feedback datasets are unable to quantitatively compare instructor and model feedback. Datasets vary in domain, feedback source, and annotation granularity (Table~\ref{tab:dataset_comparison}).
\textsc{Expos\'ia} \citep{zyska_exposi_2026} covers peer and instructor feedback on academic writing, and \citet{coyne_annotating_2025} target learner grammatical errors with an error-type and generalizability framework. \textsc{ArgRewrite v2} \citep{kashefi_argrewrite_2022} annotates revisions of argumentative essays. 
\citet{pilan_dataset_2020} examine the effect of teacher feedback characteristics on revision outcomes.
These datasets primarily focus on human feedback, but there are also datasets that instead collect and compare machine-generated writing feedback. \textsc{LEAF} \citep{behzad_leaf_2024} and its extension \textsc{LEAF++} \citep{misgna_leaf_2026} target English learners' essays with LLM feedback, with \textsc{LEAF++} adding trait-level scores. \textsc{SEFL} \citep{zhang_sefl_2026} releases synthetic formative feedback, and \citet{rashkin_help_2025} study feedback on creative story writing. \citet{mah_christopher_sentence-corrections_nodate} qualitatively compare LLM and expert teacher feedback, finding that teachers offer more dialogic, multi-level feedback while LLMs default to sentence-level corrections. \dataset complements these resources by combining sentence-anchored, parallel feedback from trained human annotators and LLMs; argumentative-goal labels; urgency rankings; and expert quality ratings atop feedback across seven dimensions (Table~\ref{tab:dataset_comparison}).
\section{The \textsc{FOXGLOVE} Dataset}

\begin{table*}[t]
\centering
\small
\begin{tabular}{lrrrrrrrr}
\toprule
Feedback giver & Total & Global & Sentence & Urgent & Comments/Essay (Median) & Mean Words (Median) \\
\midrule
Humans              & 696 & 138 & 558 & 389 & 5.0 (4.0) & 37.9 (35.0) \\
\midrule
LLM Total           & 1,644 & 276 & 1,368 & 828 & 6.0 (6.0) & 87.1 (85.0) \\[2pt]
\quad GPT-5.2             & 456 &  69 & 387 & 207 & 6.6 (7.0) & 96.6 (95.0) \\
\quad Claude Sonnet 4.5   & 503 &  69 & 434 & 207 & 7.3 (7.0) & 100.9 (94.0) \\
\quad Llama 3.3 70B       & 390 &  69 & 321 & 207 & 5.7 (6.0) & 62.8 (63.0) \\
\quad Qwen3-Next 80B      & 295 &  69 & 226 & 207 & 4.3 (4.0) & 80.9 (80.0) \\

\midrule
Total               & 2,340 & 414 & 1,926 & 1,217 & 5.7 (5.0) & 72.5 (76.0) \\
\bottomrule
\end{tabular}
\caption{Summary statistics of feedback from humans and LLMs. \textit{Sentence} denotes sentence-level feedback. \textit{Urgent} denotes feedback marked as urgent. \textit{Comments/Essay} indicates the average number of feedback comments per essay. \textit{Mean Words} denotes feedback length per comment. Table~\ref{tab:human_feedback_by_reviewer} in the Appendix shows per instructor breakdown.}
\label{tab:feedback_overview}
\end{table*}

Our dataset construction operationalizes the three properties that writing research identifies as making feedback usable and effective for revision: goal-orientation \cite{flower_cognitive_1981}, anchoring \cite{churchill_anchored_2000}, and prioritization \cite{sommers_responding_1982}. Each property addresses a distinct revision need: \emph{what}, \emph{where}, and \emph{in what order}. Feedback usable for revision requires all three. We describe the source essays (\S\ref{sec:source_essay}), the feedback schema and the literature motivating each of its components (\S\ref{sec:feedback_schema}), instructor (\S\ref{sec:human_feedback_collection}) and LLM (\S\ref{sec:llm_feedback_generation}) data collection procedures, and a descriptive analysis of \dataset (\S\ref{sec:data_description}).

\subsection{Source Essays}
\label{sec:source_essay}
We sampled essays from the PERSUADE 2.0 corpus, a large-scale corpus of argumentative essays written by high-school students \cite{crossley_large-scale_2024}. PERSUADE 2.0 annotates each essay at the sentence-level with discourse element labels that identify the argumentative roles of each sentence. We adopt these role labels as our five goal labels---Position, Claim, Evidence, Counterclaim, and Rebuttal (see definitions in \S\ref{sec:goal_definition})---to structure all feedback in the dataset. We include only essays written by grade 12 students for higher quality essays that elicit meaningful feedback beyond spelling or grammatical issues. To ensure reasonable consistency in the essay length, we removed essays in the top and bottom 5\% of the length distribution before sampling, then randomly selected 69 essays from the remaining pool. Each essay in the dataset averages 422.8 (SD = 118.5) words. 

\subsection{Feedback Schema}
\label{sec:feedback_schema}
Every feedback giver (instructor or LLM) produces feedback for each essay according to a shared schema with three components:

\begin{itemize}
    \item \textbf{Sentence-level feedback.} A free-text comment anchored to one or more contiguous sentences in the essay, addressing one of five argumentative goals. Goal definitions follow those in the PERSUADE 2.0 dataset \cite{crossley_large-scale_2024} and are adapted from human-tutor instructions used in writing-instruction practice. If a feedback giver wishes to address multiple goals on the same span of text, they produce multiple comments, one per goal.
    \item \textbf{Urgency ranking.} After producing all sentence-level comments for an essay, the feedback giver selects the three most important comments and ranks them in order of urgency for improving the essay's argumentative quality (1 = most urgent).
    \item \textbf{Global comment.} A single free-text comment per essay synthesizing how the essay aligns with the stated argumentative goals.
\end{itemize}

\subsection{Instructor Feedback Collection}
\label{sec:human_feedback_collection}
We recruited 14 U.S.-based  writing instructors with background or training in high school and college level English writing. All instructors had received explicit training programs external to our study in giving writing feedback. Instructors completed an additional paid training session (\$15, lasting around 20-30 minutes) in which they reviewed the consent document (including how their data would be used) and provided consent, then were introduced to our specific feedback schema and the goal definitions. The complete human annotation guide is included in Appendix~\ref{sec:human_instruction}. 

Each essay in the corpus was independently annotated by two instructors via separate Google Docs. Instructors worked at their own pace, using the document's commenting features to anchor feedback to specific sentences and to record goal labels, comment text, and urgency rankings (Figure~\ref{fig:real_googledoc} in the Appendix shows an example). Feedback comments violating the schema (e.g., no global comments) were returned for revision before being included in the dataset. Instructors were compensated at \$25 per hour, based on equivalent rates for writing instruction in the areas where data were collected. All personally identifiable information was removed before analysis; instructors are referred to by anonymized identifiers. The annotation protocol and data collection were approved by the relevant Institutional Review Board.

\subsection{LLM Feedback Generation}
\label{sec:llm_feedback_generation}
To complement our instructor-written feedback, we generated feedback from four LLMs: GPT-5.2, Claude Sonnet 4.5 (standard inference), Llama 3.3 70B, and Qwen3-Next 80B. Our goal was to collect feedback from multiple open and closed model families, but we did not use smaller models due to initial tests that suggested these models struggled to generate feedback within our schema (e.g., did not highlight text correctly). This aligns with prior work that has mostly used the largest available models for writing evaluations \cite{behzad_leaf_2024}. 

Each model received the same prompt template (Figure~\ref{fig:Prompt_final_feedback} in the Appendix), which mirrors the instructor instructions and requires output in a fixed JSON schema with fields for goal, sentence span, comment text, urgency rank, and global comment. The four models produced 1,368 sentence-level comments and 276 global comments. Every model response was checked against two constraints: (i)~the response must be valid JSON conforming to the schema; and (ii)~each highlighted span must consist of one or more complete and contiguous sentences in the essay. Failures were resampled over multiple rounds; a few persistent span mismatches were corrected manually. Per-model failure rates are reported in Appendix~\ref{sec:llm_instruction}.
\subsection{Description of \textsc{FOXGLOVE}}
\label{sec:data_description}

\paragraph{Dataset Overview}
Table~\ref{tab:feedback_overview} reports summary statistics, including the four models we compared. We discuss the quality ratings of both model and instructor feedback in \S\ref{sec:quality}.

\paragraph{Comment Volume and Length}

For sentence-level comments, instructors in total produced 558 comments (Mean = 4.04/essay/instructor; SD = 1.30); the four models in total produced 1,368 comments (Mean = 4.96/essay/model; SD = 0.72). Per essay, instructor comments averaged 35.53 words (SD = 11.14); model comments averaged 84.66 words (SD = 5.57). For global comments, each feedback giver was required to make one global comment per essay. The mean word count for instructor global comments is 47.88 words (SD = 15.50), and 101.34 words (SD = 6.12) for models. 

\paragraph{Goal distribution}
\label{sec:goal_distribution}
Humans and models distributed feedback across the five argumentative goals in similar proportions. \textit{Claims} received the most feedback from both groups (39.96\% instructor, 40.79\% model), followed by \textit{Position} (21.68\%) for instructor and \textit{Evidence} (20.47\%) for models. \textit{Counterclaim} (14.52\% instructor, 12.87\% model) and \textit{Rebuttal} (5.73\% instructor, 6.07\% model) were the sparsest categories for both. The similar ranking of goals suggests convergent attention to higher-level argumentative components by instructors and models.
Table~\ref{tab:goal_distribution} in the Appendix reports distributions for all goals across instructors and models.

\paragraph{Urgency Rankings}
The average number of rankings across all feedback comments is 2.94 (SD = 0.35). Across all reviewers, \textit{Claims} attracted the most urgent picks for both models and instructors (34.66\% model, 38.56\% instructors), followed by \textit{Position}, \textit{Evidence}, \textit{Counterclaim}, and \textit{Rebuttal}. Table~\ref{tab:urgent_by_goal_reviewer_type} in the Appendix includes urgency distributions by goals for instructors and models.

\begin{figure*}
    \centering
    \includegraphics[width=0.95\linewidth]{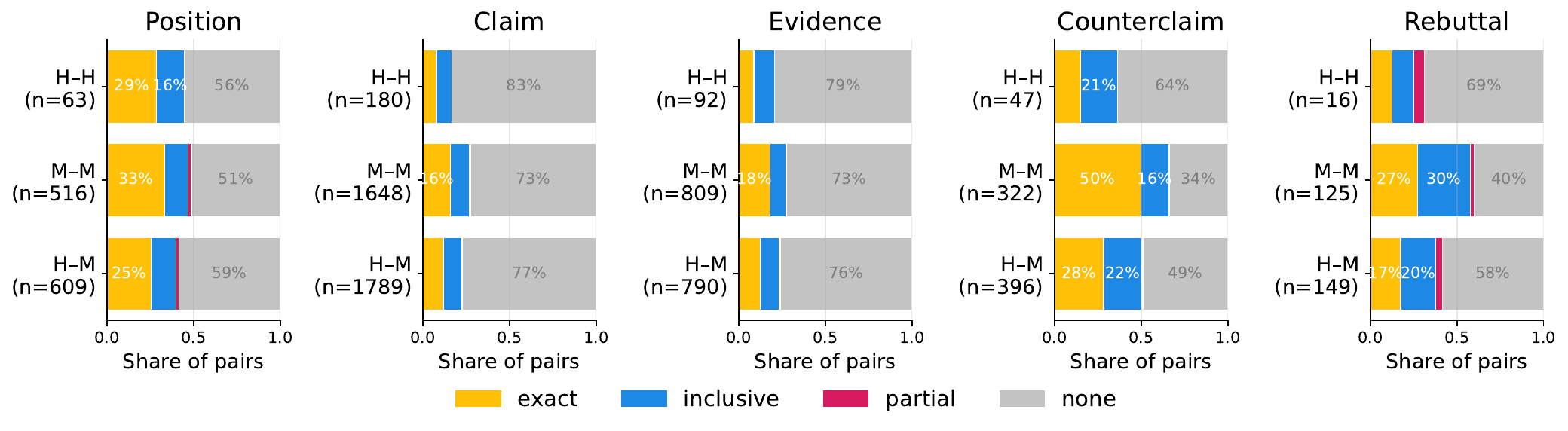}
    \caption{Sentence-span overlap between comment pairs, conditioned on both the same essay and the same argumentative goal. ``H-H'' stands for ``human-human'', ``M-M'' is ``model-model'', and ``H-M'' is ``human-model''. We found that overlap is goal-dependent, as Counterclaim and Rebuttal show far higher overlap than the others.}
    \label{fig:span_overlap}
\end{figure*}

\section{Writing Feedback Analysis}
\label{sec:analysis}

We conduct two threads of analysis on \dataset to examine differences in instructor and model comments: agreement analysis on the highlighted essay spans and language analysis on the feedback comments themselves. Specifically, we asked: 1) do LLMs and instructors highlight the same parts of an essay? and 2) how do feedback comments differ between LLMs and instructors? 

\subsection{Do LLMs and Instructors Highlight the Same Parts of an Essay?}

We compare highlight spans within instructors (Human-Human, denoted as H–H), within LLMs (Model-Model, M–M), and between instructors and LLMs (H–M) feedback pairs for each essay.
 
We first look at how much of each essay each type of feedback giver highlights. Instructors highlight on average 42.8\% of essay. However, all LLMs highlight a higher percentage of essays than instructors, with GPT-5.2 and Claude Sonnet 4.5 highlighting the majority of each essay (over 80\%). Table~\ref{tab:highlight_span_summary} in the Appendix reports the per-model highlighting behaviors. 

Since the feedback schema anchors highlights at the sentence level, we classify each comment pair as \emph{exact} agreement (both comments anchor to the same set of sentences), \emph{inclusive} (one comment's sentence set is a subset of the other's), or \emph{partial} agreement (their sentence sets intersect but neither is a subset). Overall, within-LLM pairs overlap most often (19.2\%), ahead of H–M (18.0\%) and H-H pairs (17.0\%). Among three types of overlaps, \emph{exact} overlap is the most common type for all (M-M, 62.7\%; H-H, 52.8\%; H-M, 50.7\%). 

Conditioning on the same essay \emph{and} the same goal, both instructors and models were more likely to agree on the same sentences for \textit{Counterclaim} and \textit{Rebuttal} (M-M overlap 66\% and 60\%, H-H 36\% and 31\%, respectively) than for \textit{Claim} and \textit{Evidence} (M-M 27\% each, H-H 17\% and 21\%, respectively), as shown in Figure~\ref{fig:span_overlap}. We hypothesize that counterclaim–rebuttal moves are scarce and structurally localized so the pool of candidate sentences is small and independent feedback givers converge almost by necessity. Across every goal, models tended to have the highest share of overlapping pairs, and were more likely to match the \emph{exact} span with other models. 

Among overlapping pairs where both feedback givers assigned an urgency rank, exact agreement reached only 38.5\%, with most disagreements off by a single tier (mean $|\Delta\text{rank}| = 0.79$). For all pairs, models agreed on the exact tier most often (22.2\%), but instructors were more likely to \emph{both} assign a rank at all (57.3\% vs.\ 51.4\% for M-M).

\begin{figure*}
    \centering
    \includegraphics[width=0.95\linewidth]{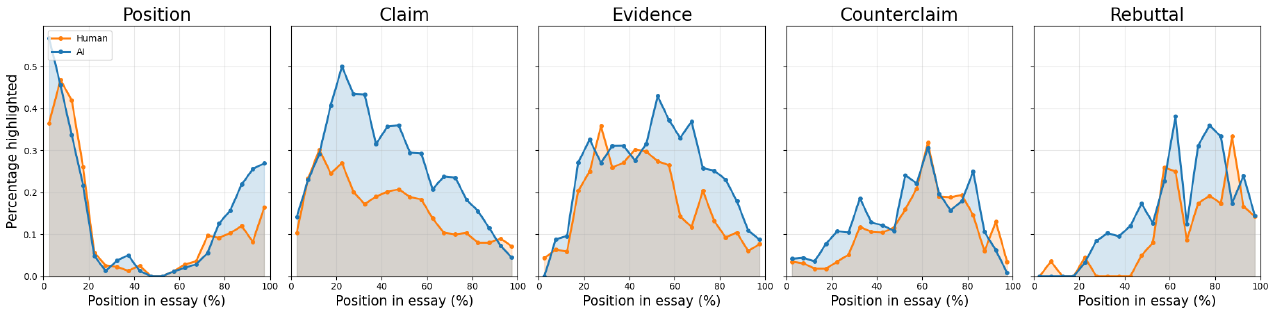}
    \caption{Highlight percentage by position for LLMs and instructor per goal. LLMs and instructor generally agree on where to highlight for each goal. The y-axis shows the percentage of sentences at a given position that are flagged for that specific goal for feedback, the x-axis shows the position in the essay in 5\% bins.}
    \label{fig:position_by_goal}
\end{figure*}

Beyond analyzing span overlaps, we ask whether instructors and LLMs attend to the same \emph{regions} of an essay overall. LLMs and instructors share similar distributions of where feedback should land within an essay (Figure~\ref{fig:position_by_goal}). For instructors and LLMs, the curves track very closely to each other for \textit{Position} and \textit{Claim}, as well as exhibit similar patterns in the rest of the categories.

Together, LLMs and instructors agree at a high level on \emph{where} each kind of feedback belongs, as their feedback follows similar positional distributions across goals. However, pairwise span overlap for models and instructors is strongly goal-dependent, high for goals like \textit{Counterclaim} and \textit{Rebuttal} but lower for \textit{Claim} and \textit{Evidence}. Even when two feedback givers do land on the same span, they diverge on \emph{how urgent} it is, agreeing on the exact urgency tier only about a fifth of the time.

\subsection{How do Feedback Comments Differ between LLMs and Instructors?}

Beyond \emph{where} feedback lands, we ask how the comments themselves differ. We highlight three differences in comment language across instructors and models: comment readability, use of personal pronouns, and use of questions. Overall, our findings are consistent with prior literature that human comments are often more personal and engaging than LLM-generated ones \cite{go-etal-2025-xdac,jianghyland2025engagement}. Full per-model results can be found in Table~\ref{tab:feedback_language_measures} in the Appendix.

\subsubsection{LLM Comments are Longer and Less Readable}
\label{sec:readability}
The readability of language impacts how easy it is for students to understand and respond to \citep{walkington2018readability, shute2008focus}. Long, complex comments may be difficult for students to act on. LLM comments are more than twice as long as instructor ones, averaging $87.1$ words against $37.9$ (Table \ref{tab:feedback_overview}), and they are markedly harder to read. When analyzing comments using Flesch--Kincaid grade level \citep{flesch1948}, a common measure of text readability, instructor feedback sits around grade $8.5$ while LLM feedback reaches $14.4$ (ranging from $13.8$ for Claude Sonnet 4.5 to $14.9$ for GPT-5.2). The Dale-Chall readability score, a complementary measure that incorporates the share of words outside a familiar-word list \citep{chall_readability_1995}, shows the same direction of complexity increase: $9.53$ for instructors compared to $11.72$ (ranging from $11.3$ for GPT-5.2 to $12.3$ for Qwen3-Next 80B) for LLMs. In practical terms, human comments read closer to a high-school level, whereas LLM comments are more demanding at roughly a college-graduate reading level. 

\subsubsection{Instructors Use More Pronouns}
\label{sec:pronoun}
Pronouns provide a compact way to measure how feedback positions the writer and the reader relationship, distinguishing feedback framed as interpersonal communication from feedback that reads as detached assessment \citep{Hyland2005Stance}. We therefore count first-person pronouns (e.g., ``I'', ``we'') and second-person pronouns (e.g., ``you'', ``your'') per $100$ words.\footnote{We first lower-cased and tokenized feedback with \texttt{spaCy}'s \texttt{en\_core\_web\_sm} model before counting pronouns.} Human feedback has a higher density of both. Humans average $1.26$ first-person pronouns per $100$ words, compared with only $0.03$ for LLMs. Human instructors also use direct address more often: $5.19$ vs.\ $3.75$ second-person pronouns per $100$ words. This contrast suggests that feedback from instructors is more conversational than that of LLMs. There is also meaningful variation across models: Claude Sonnet 4.5 and Qwen3-Next 80B approach human levels of second-person address ($5.17$ and $4.95$, respectively), while Llama 3.3 70B uses very little ($1.03$).

\subsubsection{Instructors Use More Questions}
\label{sec:questions}
Questions can involve readers by prompting them to reconsider a choice rather than simply changing texts \citep{Hyland2002WhatDT}. When analyzing question use, instructors use questions substantially more than LLMs. Similar to pronouns, we counted the number of question marks (`?') in each piece of feedback and normalized per $100$ words. This is a straightforward measure that captures explicitly punctuated questions rather than all interrogative sentences. Instructor feedback contains $0.85$ questions per $100$ words against just $0.23$ for LLMs. Among models, Claude Sonnet 4.5 is the closest to humans ($0.62$), while others rarely ask questions. 

\subsubsection{These Differences May Be Steerable}

The above differences arise under a prompt that mirrors the instructor instructions with no constraints on length or style. Having identified these differences, we can now test whether they can be steered through prompting. We re-generated feedback from all four models on all 69 essays under three additional conditions: (i) a \emph{length} instruction calibrated to the instructor average (around 40 words for sentence-level comments and 50 for global comments); (ii) a \emph{style} instruction to write conversationally at the student's level, using personal pronouns and questions; and (iii) both instructions \emph{combined}. Full prompt text and per-model results appear in Appendix~\ref{sec:steering}.

We found that stylistic features are steerable: with the style prompt, all four models moved their use of questions and second-person address to at or above human reference levels. However, length can be difficult to steer: though most models are steered to around 40 words,  Llama 3.3 70B is very sensitive to the prompt, with a mean and median feedback length of 17 words. Nevertheless, brevity alone made comments denser rather than simpler for all models (Dale–Chall 11.72$\rightarrow$12.37, averaged across models), indicating that shorter feedback is not simpler feedback. Combining both prompts shows better stylistic and length alignment than the baseline prompt alone, though it slightly increases the word count compared to only the length prompt, where the stylistic features behave differently across models when combined compared to only the style prompt (e.g., GPT-5.2 used fewer questions but others used more). 
\section{Quality Ratings of Writing Feedback}
\label{sec:quality}

\begin{table*}[t]
\centering
\small
\begin{tabular}{lccc|ccc}
\toprule
 & \multicolumn{3}{c}{Global Ratings} & \multicolumn{3}{c}{Sentence-Level Ratings} \\
\cmidrule(lr){2-4} \cmidrule(lr){5-7}
Dimension & Human & LLMs & $\Delta$ & Human & LLMs & $\Delta$ \\
\midrule

 Accuracy (position) & N/A&N/A &N/A & 4.19 (0.37) & 4.38 (0.33) &$+0.19^{*}$ \\
Accuracy (content)  & 3.80 (0.63) & 4.50 (0.24) & $+0.70^{*}$ & 3.71 (0.72) & 4.35 (0.34) & $+0.65^{***}$ \\
Actionability       & 1.95 (0.60) & 3.00 (0.53) & $+1.05^{***}$ & 2.21 (0.57) & 3.61 (0.59) & $+1.40^{***}$ \\
Clarity             & 3.00 (0.62) & 3.60 (0.52) & $+0.60^{*}$ & 2.85 (0.66) & 3.42 (0.54) & $+0.57^{***}$ \\
Relevance           & 4.30 (0.35) & 4.50 (0.24) & $+0.20$ & 3.93 (0.70) & 4.35 (0.31) & $+0.43^{***}$ \\
Specificity         & 2.50 (0.85) & 4.15 (0.58) & $+1.65^{***}$ & 1.96 (0.48) & 3.80 (0.66) & $+1.85^{***}$ \\
Tone                & 3.95 (0.69) & 3.20 (0.35) & $-0.75^{**}$ & 3.35 (0.69) & 3.17 (0.33) & $-0.19$ \\
\midrule

\end{tabular}
\caption{Human ratings of feedback quality. Values are means (SD) on a 1--5 scale; $\Delta$ denotes LLMs $-$ Human. Positional accuracy is not relevant to global feedback ratings. Significance of each $\Delta$ is from a linear
mixed-effects model with Holm correction ($^{*}p<.05$; $^{**}p<.01$; $^{***}p<.001$; unmarked differences are not significant). Full estimates with standard errors appear in Table~\ref{tab:mixed-effects}. }
\label{tab:human_ratings}
\end{table*}

To complement our analysis of feedback content, we explore how writing instructors perceive the quality of instructor-written and model-generated writing feedback. Prior work has generally characterized model-generated writing feedback as improving rapidly with overall model improvement: work on earlier models (e.g., GPT-4) suggested that generated feedback often focused on surface level characteristics of texts \citep{liang_can_2023}, while more recent work has found model feedback on students' works to be broadly comparable to that of human experts \cite{steiss_comparing_2024, lu_ai-mediated_2026}. Beyond the agreement and content analysis in \S\ref{sec:analysis}, we collect and analyze expert instructor quality ratings to assess how feedback from each source compares along dimensions important for writing instruction. 

\subsection{Expert Rating Collection Process}

We recruited two additional college-level English instructors at a U.S.-based university as our expert raters. We selected seven dimensions to rate feedback on to capture: informational quality of feedback (Accuracy of content, Accuracy of position, Relevance, Specificity), feedback usability for revision (Actionability, Clarity), and interpersonal aspects (Tone) \citep{steiss_comparing_2024, behzad_assessing_2024, rashkin_help_2025, liu_crafting_2026}. Global feedback was not rated on positional accuracy. The two raters first familiarized themselves with the rating dimensions and interface (see Figure~\ref{fig:interface} in the Appendix) by rating four sets of feedback for one essay and calibrating ratings in a meeting with two researchers. We revised guidelines for the rating dimensions based on discussions with raters during the practice round (Appendix~\ref{sec:rating} has more details). We did not include the data collected in this practice round. Raters worked independently and were blind to the source of each comment.

In total, we collected 1{,}430 ratings from 105 comments (61 from LLMs, 44 from humans) across five essays randomly sampled from the corpus. For each essay we showed raters feedback from the two human instructors and two randomly sampled models. Each comment was then rated on 5-point Likert scales across the seven dimensions. The inter-rater reliability (IRR) between the two raters was Cohen's $\kappa_w$ = 0.36 for sentence-level feedback and $\kappa_w$ = 0.35 for global feedback, indicating a fair level of agreement for a subjective task like feedback quality rating \citep{landis1977measurement, sap2020socialbiasframes, da-san-martino-etal-2019-fine}. Adjacent (±1) agreement was 78.8\% for sentence-level and 76.7\% for global feedback, indicating that most rating disagreements fell within a single scale point.

\subsubsection{LLM Feedback Receives Higher Ratings, Consistent with a Substantial Length Effect}

We report the rating statistics for all comments in Table~\ref{tab:human_ratings}. For sentence level feedback, LLM feedback scored higher than human feedback on six out of seven dimensions, with the highest difference being in specificity ($\Delta = +1.85$), followed by actionability ($+1.40$). The only dimension models did not perform better on was tone ($-0.19$). The pattern for global comments was similar. One interesting pattern we observed was that our two raters often cited feedback being more comprehensive as their reasons for higher ratings. Because LLM feedback was substantially longer than human feedback (see Table~\ref{tab:feedback_overview}), we further analyzed how comment length was associated with quality ratings. 

To identify quality differences driven by length, we computed length-adjusted means for each dimension \citep{chodorow_beyond_2004, dubois_length-controlled_2025}. For a dimension $D$, we fit an ordinary least square regression controlled for length, then evaluate at the overall mean length ($\approx 64$ words) and add back each side's mean residual. This yields the human and LLM means we would expect if both produced feedback of average length. We refer to these estimates as adjusted scores.

We found that a substantial part of the apparent human-model rating gap appears to be driven by length. Adjustment converges \emph{both} sides rather than just lowering LLM scores: at the average length, human ratings rise and LLM ratings fall on every dimension except tone. The effect is largest for specificity ($1.85\!\to\!0.45$) and actionability  ($1.40\!\to\!0.42$) for both human and model. We report the full analysis in Appendix \ref{sec:length_quality}.

\section{Discussion}

\subsection{What Length Measures}

The length difference between human and LLM feedback raises a broader question about what counts as ``quality'' in formative feedback. One way to interpret our finding is that longer comments simply look more thorough for human raters, thereby giving the appearance of ``good'' feedback \citep{bu-etal-2025-beyond}. Another possible interpretation is that longer comments have a higher likelihood of containing good, specific, and actionable feedback. This creates a tension, as LLMs can generate long feedback cheaply while humans cannot. While longer feedback may be considered higher quality in controlled settings, it may not lead to meaningful revision \citep{wu_feedback_2020}. In fact, writers may simply ignore or skim longer feedback comments \citep{shute2008focus}. Future use of \dataset should examine whether longer LLM feedback translates into stronger student revisions, comprehension, and engagement, complementing expert ratings with downstream student outcomes.

\subsection{The Value of Variation}
An open question our dataset raises is the value of \emph{variation} in feedback. We observe greater variation among human feedback than among LLMs, which tend to converge on similar spans (e.g., span overlap is higher between models than between instructors or between instructors and models). Whether this consistency is desirable can be interpreted both ways. Variation can be a strength: good feedback is tailored to a particular writer, goal, and moment \citep{mah_christopher_sentence-corrections_nodate}, so the homogenization of feedback across models may flatten the reader-specific responsiveness that makes human feedback effective. But variation is not automatically good, as it can also reflect inconsistency or noise. 

This question can be interesting to both audiences \dataset is meant to serve. For \textbf{model evaluators}, it reframes what alignment means: rather than optimizing solely toward expert-rated quality, \dataset can serve as a testbed for whether eliciting more varied, reader-specific feedback improves usefulness. For \textbf{writing instructors}, it informs when LLM feedback is a reasonable substitute for, or complement to, human feedback, and where it falls short. Because \dataset pairs feedback with quality ratings and highlighted spans along pedagogical dimensions (e.g., specificity, actionability, tone), it offers a reference point for judging whether a model's consistency is an asset or a liability in a given classroom use.

\subsection{What \textsc{FOXGLOVE} enables}

\paragraph{Studying feedback uptake and revision}

\dataset can serve as a starting point for controlled studies of how students act on feedback. Future work could have students revise essays based on different feedback, connecting expert ratings to outcomes such as revision quality, comprehension, and perceived usefulness---closing the loop between feedback quality as rated and as experienced.

\paragraph{Reusable protocols}

Our annotation protocol is designed for reuse. We include the full protocols for instructor (Appendix~\ref{sec:human_instruction}) and LLM (Appendix~\ref{sec:llm_instruction}) feedback collection, and quality rating (Appendix~\ref{sec:rating}). We release all code for LLM feedback generation and the rater interface\footnote{\url{https://github.com/yijunliu23/foxglove_data_release}} . We hope this lowers the cost of benchmarking future systems against a common baseline, and we invite extensions of \dataset{} to additional models as capabilities evolve.

\section{Conclusion}

We introduce \dataset, a dataset of 696 feedback comments from trained human instructors paired with 1,644 comments from four LLMs, totaling 2,340 feedback comments across 69 twelfth-grade argumentative essays, paired with 1,430 expert quality ratings on a subset of feedback. Our analyses show that humans and LLMs agree broadly on where feedback belongs across argumentative goals and essay positions, but diverge at the sentence level and on urgency ratings. While LLM feedback receives higher expert ratings on six of seven quality dimensions, this advantage may come from comment length, as LLMs are more likely to generate substantially longer comments. \dataset provides reusable protocols and a common benchmark to support that next step such as connecting rated quality to revision outcomes. 

\section*{Limitations}

We study a specific form of feedback: comments tied to predefined argumentative goals and assigned an urgency rank, collected in an academic setting at both the sentence and global level. This schema makes feedback comparable and analyzable, but it also privileges feedback organized around these predefined argumentative moves and may underrepresent feedback that falls outside it (e.g., encouragement and praise). Because both human and model feedback were collected under this schema, our results characterize feedback \emph{within this paradigm} and may draw attention disproportionately to the types of feedback the schema makes easy to express.

The feedback in \dataset was collected for the purpose of this study rather than drawn from authentic instructional settings. Feedback givers wrote feedback on essays they were assigned, without an ongoing relationship with the writer or knowledge of the writer's history. As a result, our findings describe how humans produce feedback \emph{under these elicitation conditions}.

\section*{Ethical Considerations}
Building on the first limitation, our schema and rubric encode one particular view of what good feedback is. Therefore, we intend \dataset as one lens on feedback quality, not a definitive standard, and we encourage its use alongside other measures rather than as a sole criterion.

Additionally, our analysis compares a fixed set of models, raters, and essays at one point in time, and our quality ratings reflect the judgments of a small number of expert raters. These results should be read as one data point. In particular, we caution against using our finding that LLM feedback receives higher ratings to justify replacing human instructors: our ratings do not directly measure what actually helps students revise. Decisions about whether and how to deploy LLM feedback in education should rest on downstream student experience and outcomes, not on rated quality alone.

\newpage

\bibliography{custom}
\twocolumn
\appendix
\section{Appendix}
\label{sec:appendix}

The PERSUADE corpus \cite{crossley_large-scale_2024} is distributed under a CC BY-NC-SA 4.0 license.\footnote{\url{https://creativecommons.org/licenses/by-nc-sa/4.0/}} Our use is consistent with its intended research use, and we release our derived data under the same CC BY-NC-SA 4.0 license to comply with its ShareAlike terms. All artifacts are intended for research use only.

\subsection{Human Feedback Instruction}
\label{sec:human_instruction}

Below is the instruction document we sent to instructors. Each essay is paired with three goals. We selected these by first identifying the argumentative goals that PERSUADE 2.0 did not rate as fully achieved for that essay, then randomly sampling three of them. LLMs and humans receive the same set of goals for each essay. 

You may provide one or multiple feedback comments addressing one or multiple goals, depending on where you see the need for improvement. Before starting to provide feedback, please first read through the essay. 

\paragraph{For each goal:}

\begin{itemize}
    \item Identify the relevant part(s) of the essay (specific sentences or sections).
    \item Each feedback should only be addressing one goal; if you need to address multiple goals for the same selection of sentences, write multiple feedback comments.
    \item When selecting a section to give feedback, the section should at least be a continuous unit of full sentences (e.g., you should not select a half sentence).
    \item Write feedback that:
    \begin{itemize}
        \item Addresses the issue described in the goal.
        \item Explains what the problem is and why it matters.
        \item Suggests ways to improve without rewriting the student’s text.
        \item Do NOT address lower order concerns such as grammatical, punctuation, spelling, or sentence structure. 
        \item Use different comment threads for different feedback. 
        \item The feedback should be standalone (e.g., do not refer to earlier feedback without details). 
        \item Maintain a constructive tone. Encourage revision and reflection rather than correction.
        \item Write in complete sentences with clear grammar and style.
    \end{itemize}
\end{itemize}

\paragraph{After Writing Feedback}
Select and then rank the top three comments that are most urgent for improving the essay’s argumentative quality. To rank, reply directly to each selected comment with its rank number 
(1 = most urgent, then 2, and 3). Write as Rank 1/2/3.

To Rank: 1) Hover on your feedback comment 2) Click on your feedback comment, a text box will appear; put in your rank 3) Click “Reply”. 

\subsubsection{Goal Definition}
\label{sec:goal_definition}
\begin{itemize}
    \item Thesis (Position): State a clear, specific stance that's closely related to the topic of the prompt.
    \item Claim: Build claims that are closely relevant to the writer’s position and back it up with specific points or perspectives.
    \item Counterclaim: Present a reasonable, relevant objection that represents a valid opposing view to the writer’s position.
    \item Rebuttal: Directly answer and refute the counterclaim.
    \item Evidence: Provide evidence that's closely relevant to the writer’s claims and backs them up with concrete facts, examples, research, statistics, or studies.
\end{itemize}

\begin{figure*}
    \centering
    \includegraphics[width=1\linewidth]{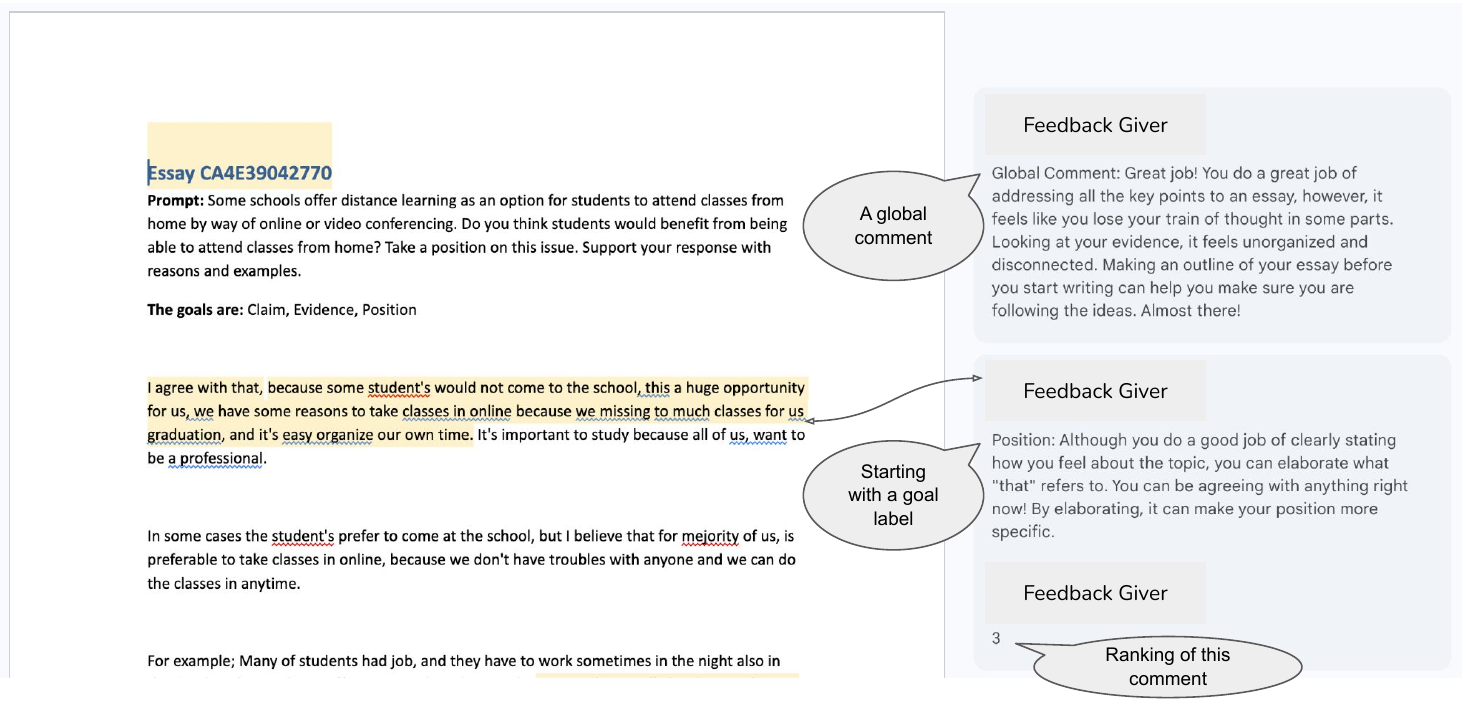}
    \caption{The Feedback Giver interface on Google Docs, showing global and span-level comments on a student essay, each with a goal label and a priority ranking.}
    \label{fig:real_googledoc}
\end{figure*}

\subsection{LLM Instructions and Output Validation}
\label{sec:llm_instruction}

\raggedbottom

We prompted each of four models (\texttt{claude-sonnet-4-5},
\texttt{gpt-5.2}, \texttt{llama3-3-70b}, and
\texttt{qwen3-next-80b}) to generate feedback for every essay
in our dataset using the template shown in Figure~\ref{fig:Prompt_final_feedback} at temperature 0. 
\texttt{gpt-5.2} was accessed through the official OpenAI API, while all other models were accessed through the Amazon Bedrock API.

\paragraph{Retry procedure.}
Each generated record was checked against the feedback schema described in \S\ref{sec:feedback_schema}. Records that failed any check were re-generated. Round $v_0$ denotes the initial generation; in round $v_{k}\,(k\geq 1)$, every (essay, model) pair that still had at least one failing run in $v_{k-1}$ was re-sampled with three fresh attempts at temperature $0$. The loop continued until either all records validated or no further progress was made. In our experiments, \texttt{claude-sonnet-4-5} and
\texttt{llama3-3-70b} converged automatically (by $v_2$ and $v_4$ respectively), but \texttt{gpt-5.2} and \texttt{qwen3-next-80b} plateaued: through $v_8$, both models continued to fail on the same handful of span-highlighting cases, regenerating slightly paraphrased quotes rather than verbatim text on every retry. For these residual records we therefore performed a manual intervention at $v_9$, editing the offending highlight strings by hand to match the essay verbatim. Table~\ref{tab:rounds} reports the per-round failure rate, computed as the fraction of records \emph{attempted in that round} that did not pass validation; the $v_9$ column for \texttt{gpt-5.2} and \texttt{qwen3-next-80b} reflects this manual correction. Dashes (--) indicate that the model had already fully converged before that round.

\paragraph{Failure modes.}
We classify each failure by the first violated criterion. Table~\ref{tab:failures} aggregates failure counts across all rounds. The categories are: \emph{generation\_error}---the response could not be JSON-extracted at all and was logged as a raw string; \emph{span\_not\_in\_essay}---a feedback item quoted text that did not match any region of the essay; and \emph{urgency\_rank}---the \texttt{Urgency\_Rank} field was empty or repeated. Notably, all generations included a global comment, so there were no failures in this category at all.

\begin{table*}[t]
\centering
\small
\setlength{\tabcolsep}{6pt}
\renewcommand{\arraystretch}{1.15}
\begin{tabular}{l rrrrrrrrrr}
\toprule
\textbf{Model} & $v_0$ & $v_1$ & $v_2$ & $v_3$ & $v_4$ & $v_5$ & $v_6$ & $v_7$ & $v_8$ & $v_9$ \\
\midrule
claude-sonnet-4-5 & 0.013 & 0.167 & 0.000 & --    & --    & --    & --    & --    & --    & --    \\
gpt-5.2           & 0.013 & 0.143 & 0.333 & 0.333 & 0.500 & 0.333 & 0.333 & 0.333 & 0.333 & 0.000 \\
llama3-3-70b      & 0.026 & 0.417 & 0.333 & 0.167 & 0.000 & -- & --    & --    & --    & --    \\
qwen3-next-80b    & 0.094 & 0.750 & 0.333 & 0.667 & 0.750 & 0.667 & 0.778 & 0.556 & 0.556 & 0.000 \\
\bottomrule
\end{tabular}
\caption{Per-round failure rate by model. Dashes indicate rounds not reached.}
\label{tab:rounds}
\end{table*}

\begin{table*}[t]
\centering
\small
\setlength{\tabcolsep}{8pt}
\renewcommand{\arraystretch}{1.15}
\begin{tabular}{l rrr r}
\toprule
\textbf{Model} & \textbf{Generation} & \textbf{Span Not} & \textbf{Urgency} & \textbf{Total} \\
               & \textbf{Error}      & \textbf{in Essay} & \textbf{Rank}    &                \\
\midrule
qwen3-next-80b    & 0 & 65 & 24 & 89 \\
gpt-5.2           & 0 & 21 &  0 & 21 \\
llama3-3-70b      & 0 & 15 &  0 & 15 \\
claude-sonnet-4-5 & 4 &  0 &  0 &  4 \\
\bottomrule
\end{tabular}
\caption{Failure counts by category, aggregated across all retry rounds.}
\label{tab:failures}
\end{table*}

\begin{figure*}[t]
    \centering
    \begin{tcolorbox}[
    taggingPrompt,
    title={\textbf{Prompt for Feedback Generation}},
    width=\textwidth
    ]
\small
\renewcommand{\arraystretch}{1.3}

You are an expert high school writing instructor. Provide formative, constructive feedback on the following student essay based on three specific writing goals. \\

**Goal Definitions:**
- Position: State a clear, specific stance that's closely related to the topic of the prompt.
- Claim: Build claims that are closely relevant to the writer’s position and back it up with specific points or perspectives.
- Counterclaim: Present a reasonable, relevant objection that represents a valid opposing view to the writer’s position.
- Rebuttal: Directly answer and refute the counterclaim.
- Evidence: Provide evidence that's closely relevant to the writer’s claims and backs them up with concrete facts, examples, research, statistics, or studies. \\

**Instructions:**
Address only the issues reflected in the assigned goals. You may provide one or multiple feedback comments addressing one or multiple goals, depending on where you see the need for improvement.\\

For each goal:
1. Identify the relevant part(s) of the essay (specific sentences or sections verbatim in the essay). When selecting a section to give feedback, the section should at least be a continuous unit of full sentences (e.g., you should not select a half sentence).
2. Each feedback should only address one goal; if you need to address multiple goals for the same selection of sentences, write multiple feedback comments.
3. Write feedback that: a) Addresses the issue described in the goal; b) Explains what the problem is and why it matters; c) Suggests ways to improve without rewriting the student’s text; d) Do NOT address lower order concerns such as grammatical, punctuation, spelling, or sentence structure; e) The feedback should be standalone (e.g., do not refer to earlier feedback without details). 
3. Maintain a constructive tone. Encourage revision and reflection rather than correction.
4. Write in complete sentences with clear grammar and style.\\

After providing feedback to each goal:
1. Rank the top three individual comments that are most urgent for improving the essay’s argumentative quality (1 = most urgent). Each rank should apply to one comment only, not an entire goal. One goal can have multiple comments; each comment is ranked individually. You should only rank the top three comments; other comments should have "None" as their rank.
2. Provide a global comment that synthesizes the issues in the essay. \\

Return your response in JSON form.\\

Essay:
\{essay\_text\} \\

Three Goals:
\{goals\} \\

JSON Format:

\begin{verbatim}
{{
    "Feedback": [
        {{
            "Goal_addressed": "",
            "Sentences/sections": "",
            "Feedback_content": "",
            "Urgency_Rank": "1, 2, 3, or None"
        }}
    ],
    "Global_Comment": "",

}}
\end{verbatim}

    \end{tcolorbox}
    \caption{Prompt for LLM Feedback Generation adapted from human tutor instructions.}
    \label{fig:Prompt_final_feedback}
\end{figure*}

\subsection{Additional Description and Analysis Results }

\begin{table}[!h]
\centering
\small
\begin{tabular}{lrrrr}
\toprule
 & \multicolumn{2}{c}{Characters} & \multicolumn{2}{c}{\% Highlighted} \\
\cmidrule(lr){2-3} \cmidrule(lr){4-5}
Feedback giver & Mean & Median & Mean & Median \\
\midrule
Instructor               &  931 &  845 & 42.8\% & 36.8\% \\
\midrule
GPT-5.2             & 1897 & 1778 & 88.3\% & 83.2\% \\
Claude Sonnet 4.5   & 1798 & 1712 & 82.1\% & 78.8\% \\
Llama 3.3 70B       & 1153 & 1069 & 53.8\% & 49.7\% \\
Qwen3-Next 80B      & 1010 &  963 & 46.2\% & 39.5\% \\
\bottomrule
\end{tabular}
\caption{Summary statistics of highlighted spans per (essay, feedback giver). For humans, values are averaged across the two feedback givers per essay.}
\label{tab:highlight_span_summary}
\end{table}

\begin{table*}[t]
\centering
\small
\setlength{\tabcolsep}{8pt}
\renewcommand{\arraystretch}{1.15}
\begin{tabular}{lrrrrrr}
\toprule
Feedback Giver & Total & Global & Sentence & Urgent & Comments/Essay (Median) & Mean Words (Median) \\
\midrule
Human (all) & 696 & 138 & 558 & 389 & 5.04 (4) & 37.94 (35.0) \\
\quad R1    &  36 &   9 &  27 &  24 & 4.00 (4) & 35.25 (35.5) \\
\quad R2    &  32 &   7 &  25 &  21 & 4.57 (5) & 31.62 (28.0) \\
\quad R3    &  10 &   2 &   8 &   3 & 5.00 (5) & 35.70 (31.0) \\
\quad R4    &  57 &  10 &  47 &  29 & 5.70 (5) & 61.51 (60.0) \\
\quad R5    &  70 &  17 &  53 &  51 & 4.12 (4) & 24.33 (20.0) \\
\quad R6    &  74 &   9 &  65 &  26 & 8.22 (8) & 18.51 (13.5) \\
\quad R7    &  42 &   8 &  34 &  23 & 5.25 (5) & 36.14 (34.0) \\
\quad R8    &  81 &  20 &  61 &  60 & 4.05 (4) & 35.89 (36.0) \\
\quad R9   &  31 &   5 &  26 &   9 & 6.20 (6) & 36.19 (34.0) \\
\quad R10   &  70 &  15 &  55 &  45 & 4.67 (4) & 38.06 (37.5) \\
\quad R11   &  30 &   9 &  21 &  17 & 3.33 (3) & 20.87 (20.5) \\
\quad R12   &  27 &   6 &  21 &  18 & 4.50 (4) & 42.22 (43.0) \\
\quad R13   &  85 &  12 &  73 &  36 & 7.08 (7) & 62.05 (61.0) \\
\quad R14   &  51 &   9 &  42 &  27 & 5.67 (5) & 38.02 (38.0) \\
\bottomrule
\end{tabular}
\caption{Summary statistics of human feedback by each individual instructor feedback giver.}
\label{tab:human_feedback_by_reviewer}
\end{table*}

\begin{table*}[t]
\centering
\small
\setlength{\tabcolsep}{8pt}
\renewcommand{\arraystretch}{1.15}
\begin{tabular}{lrrrrrr}
\toprule
Feedback Giver & Position & Claim & Evidence & Counterclaim & Rebuttal & Row Total \\
\midrule
Humans & 121 (21.7\%) & 223 (40.0\%) & 101 (18.1\%) & 81 (14.5\%) & 32 (5.7\%) & 558 \\
\midrule
LLM Total & 271 (19.8\%) & 558 (40.8\%) & 280 (20.5\%) & 176 (12.9\%) & 83 (6.1\%) & 1368 \\
\quad GPT-5.2           & 87 (22.5\%) & 161 (41.6\%) & 75 (19.4\%) & 45 (11.6\%) & 19 (4.9\%) & 387 \\
\quad Claude Sonnet 4.5 & 76 (17.5\%) & 197 (45.4\%) & 95 (21.9\%) & 47 (10.8\%) & 19 (4.4\%) & 434 \\
\quad Llama 3.3 70B     & 56 (17.4\%) & 121 (37.7\%) & 70 (21.8\%) & 48 (15.0\%) & 26 (8.1\%) & 321 \\
\quad Qwen3-Next 80B    & 52 (23.0\%) & 79 (35.0\%)  & 40 (17.7\%) & 36 (15.9\%) & 19 (8.4\%) & 226 \\
\midrule
Total & 392 (20.4\%) & 781 (40.6\%) & 381 (19.8\%) & 257 (13.3\%) & 115 (6.0\%) & 1926 \\
\bottomrule
\end{tabular}
\caption{Distribution of feedback comments across argumentative goals by feedback giver.}
\label{tab:goal_distribution}
\end{table*}

\begin{table*}[htbp]
\centering
\small
\begin{tabular}{l r r r r r r}
\toprule
\textbf{Feedback giver} & \textbf{Position} & \textbf{Claim} & \textbf{Evidence} & \textbf{Counterclaim} & \textbf{Rebuttal} & \textbf{Row total} \\
\midrule
Humans            & 88 (22.6\%)  & 150 (38.6\%) & 65 (16.7\%)  & 59 (15.2\%)  & 27 (6.9\%) & 389 \\
LLM Total         & 198 (23.9\%) & 287 (34.7\%) & 146 (17.6\%) & 137 (16.5\%) & 60 (7.2\%) & 828 \\
\quad GPT-5.2           & 60 (29.0\%) & 65 (31.4\%) & 36 (17.4\%) & 34 (16.4\%) & 12 (5.8\%) & 207 \\
\quad Claude Sonnet 4.5 & 49 (23.7\%) & 76 (36.7\%) & 36 (17.4\%) & 30 (14.5\%) & 16 (7.7\%) & 207 \\
\quad Llama 3.3 70B     & 42 (20.3\%) & 73 (35.3\%) & 39 (18.8\%) & 38 (18.4\%) & 15 (7.2\%) & 207 \\
\quad Qwen3-Next 80B    & 47 (22.7\%) & 73 (35.3\%) & 35 (16.9\%) & 35 (16.9\%) & 17 (8.2\%) & 207 \\
\midrule
Total             & 286 (23.5\%) & 437 (35.9\%) & 211 (17.3\%) & 196 (16.1\%) & 87 (7.1\%) & 1217 \\
\bottomrule
\end{tabular}
\caption{Distribution of urgent feedback comments across argumentative goals by feedback giver. }
\label{tab:urgent_by_goal_reviewer_type}
\end{table*}

\begin{table*}[t]
\centering
\small
\begin{tabular}{l|cc|cc|c}
\toprule
\multirow{2}{*}{Feedback giver} & \multicolumn{2}{c|}{Readability} & \multicolumn{2}{c|}{Pronouns (per 100 words)} & \multirow{2}{*}{\makecell{Questions\\(per 100 words)}}  \\[1pt]
& F-K grade & Dale--Chall & First-person & Second-person & \\
\midrule
Instructors
& $8.53 \pm 2.93$
& $9.53 \pm 1.71$
& $1.26 \pm 2.33$
& $5.19 \pm 4.83$
& $0.85 \pm 2.67$ \\
\midrule
LLM Average
& $14.37 \pm 2.67$
& $11.72 \pm 1.00$
& $0.03 \pm 0.22$
& $3.75 \pm 2.71$
& $0.23 \pm 0.71$ \\[2pt]
\quad GPT-5.2
& $14.92 \pm 3.01$
& $11.29 \pm 0.92$
& $0.04 \pm 0.22$
& $3.74 \pm 2.19$
& $0.05 \pm 0.26$ \\
\quad Claude Sonnet 4.5
& $13.80 \pm 1.96$
& $11.77 \pm 0.89$
& $0.04 \pm 0.23$
& $5.17 \pm 2.52$
& $0.62 \pm 1.08$ \\
\quad Llama 3.3 70B
& $14.18 \pm 3.05$
& $11.73 \pm 1.05$
& $0.02 \pm 0.20$
& $1.03 \pm 1.96$
& $0.03 \pm 0.24$ \\
\quad Qwen3-Next 80B
& $14.77 \pm 2.42$
& $12.32 \pm 0.93$
& $0.04 \pm 0.22$
& $4.95 \pm 1.75$
& $0.09 \pm 0.55$ \\
\bottomrule
\end{tabular}
\caption{Language and content measures of feedback from humans and LLMs. Each cell reports the mean $\pm$ standard deviation across feedback comments. F-K grade denotes Flesch--Kincaid grade level. Dale--Chall denotes the Dale--Chall readability score. Higher values for both readability measures indicate less readable text. First-person and second-person pronouns are reported per 100 words. Questions are measured as question marks per 100 words.}
\label{tab:feedback_language_measures}
\end{table*}

\begin{figure*}
    \centering
    \includegraphics[width=0.95\linewidth]{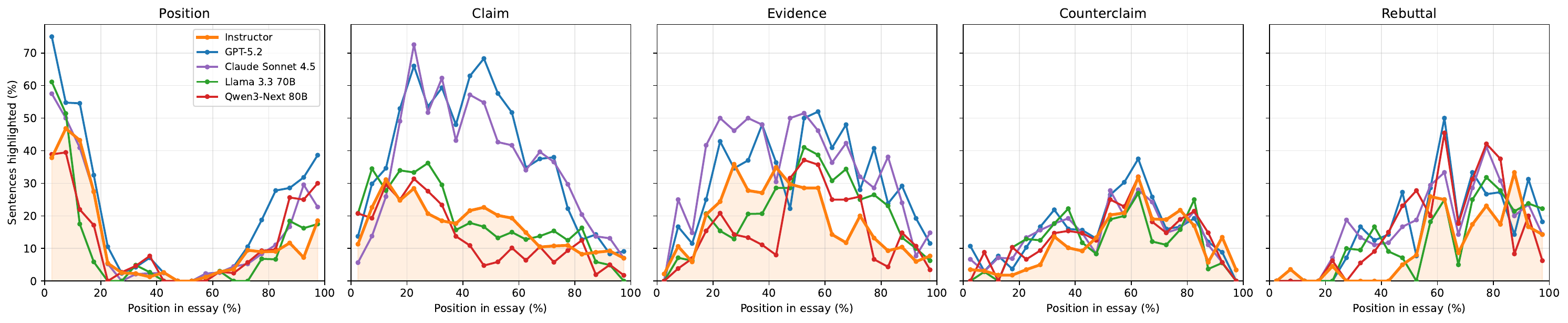}
    \caption{Highlight percentage by position for LLMs and instructor per goal, shown individually for each model. The y-axis shows the percentage of sentences at a given position that are flagged for that specific goal for feedback, the x-axis shows the position in the essay in 5\% bins.}
    \label{fig:position_by_goal_by_model}
\end{figure*}

\subsection{LLM Steering Prompts}
\label{sec:steering}

We appended the following additional prompt to the original prompt (Figure~\ref{fig:Prompt_final_feedback}) for length and style steering. 

\paragraph{Length instruction.}
\begin{quote}\itshape
Each sentence-level feedback comment is suggested to be around 40 words. The
global comment is suggested to be around 50 words.
\end{quote}

\paragraph{Style instruction.}
\begin{quote}\itshape
Write in a conversational tone, as if speaking directly with the student, and
give feedback in a way that is coherent to the student's writing level.
Address the writer as ``you,'' use first person (``I'') where natural, and
pose questions that invite the writer to reconsider their choices rather than
only giving directives.
\end{quote}

\paragraph{Combined condition.} Both instructions above are appended to the
default prompt.

\begin{table*}[t]
\centering
\small
\setlength{\tabcolsep}{8pt}
\renewcommand{\arraystretch}{1.15}
\begin{tabular}{l|cc|cc|cc|c}
\toprule
\multirow{2}{*}{Prompt} & \multicolumn{2}{c|}{Length (words)} & \multicolumn{2}{c|}{Readability} & \multicolumn{2}{c|}{Pronouns (per 100 words)} & \multirow{2}{*}{\makecell{Questions\\(per 100 words)}}  \\[1pt]
 & Global & Sentence & F-K grade & Dale--Chall & First & Second & \\
\midrule
\multicolumn{8}{l}{\textit{LLM Total}} \\[1pt]
\quad Baseline & 101.3 & 84.2 & 14.37 & 11.72 & 0.03 & 3.75 & 0.23 \\
\quad Length & 46.9 & 41.2 & 12.99 & 12.37 & 0.02 & 3.74 & 0.19 \\
\quad Style & 116.0 & 88.8 & 10.98 & 10.09 & 0.91 & 6.96 & 1.58 \\
\quad Both & 51.1 & 44.0 & 10.68 & 10.76 & 0.43 & 7.42 & 1.97 \\
\midrule
\multicolumn{8}{l}{\textit{GPT-5.2}} \\[1pt]
\quad Baseline & 93.1 & 97.3 & 14.92 & 11.29 & 0.04 & 3.74 & 0.05 \\
\quad Length & 53.3 & 46.1 & 12.40 & 11.60 & 0.04 & 3.52 & 0.01 \\
\quad Style & 109.3 & 95.6 & 11.87 & 10.31 & 0.19 & 5.93 & 0.89 \\
\quad Both & 53.3 & 45.5 & 10.58 & 10.59 & 0.15 & 5.22 & 0.82 \\
\midrule
\multicolumn{8}{l}{\textit{Claude Sonnet 4.5}} \\[1pt]
\quad Baseline & 159.4 & 91.6 & 13.80 & 11.77 & 0.04 & 5.17 & 0.62 \\
\quad Length & 63.5 & 50.0 & 13.50 & 12.53 & 0.01 & 4.77 & 0.50 \\
\quad Style & 165.6 & 93.9 & 10.73 & 10.10 & 0.81 & 7.17 & 1.98 \\
\quad Both & 67.9 & 54.7 & 10.69 & 10.86 & 0.40 & 6.74 & 2.35 \\
\midrule
\multicolumn{8}{l}{\textit{Llama 3.3 70B}} \\[1pt]
\quad Baseline & 73.1 & 60.6 & 14.18 & 11.73 & 0.02 & 1.03 & 0.03 \\
\quad Length & 21.7 & 17.2 & 13.09 & 12.82 & 0.00 & 0.84 & 0.02 \\
\quad Style & 89.2 & 72.8 & 11.38 & 9.96 & 1.90 & 8.01 & 1.69 \\
\quad Both & 23.6 & 20.3 & 12.13 & 11.14 & 0.15 & 10.83 & 2.79 \\
\midrule
\multicolumn{8}{l}{\textit{Qwen3-Next 80B}} \\[1pt]
\quad Baseline & 79.8 & 81.3 & 14.77 & 12.32 & 0.04 & 4.95 & 0.09 \\
\quad Length & 49.4 & 44.0 & 12.94 & 12.81 & 0.01 & 5.61 & 0.16 \\
\quad Style & 99.8 & 88.3 & 9.43 & 9.86 & 0.98 & 6.94 & 1.85 \\
\quad Both & 59.5 & 49.4 & 9.13 & 10.45 & 1.22 & 8.17 & 2.19 \\
\bottomrule
\end{tabular}
\caption{Language and content measures of LLM feedback under the baseline prompt and prompts tuned for length, style, or both (Both), grouped by model. Each cell reports the mean across feedback comments. Length is the number of words per comment, reported separately for global (essay-level) and sentence-level feedback. F-K grade denotes Flesch--Kincaid grade level. Dale--Chall denotes the Dale--Chall readability score. Higher values for both readability measures indicate less readable text. First-person and second-person pronouns are reported per 100 words. Questions are measured as question marks per 100 words.}
\label{tab:feedback_language_measures_prompt_tuned}
\end{table*}

\subsection{Human Rater Instruction}
\label{sec:rating}

\begin{enumerate}
    \item \textbf{How relevant is this feedback to the assigned writing goal?} \\
    \textit{Not at all $\cdot$ Slightly $\cdot$ Moderately $\cdot$ Highly $\cdot$ Fully} \\
    We want to know if the feedback stays on topic with the assigned writing goal. If the feedback includes other irrelevant or less-relevant topics, you can lower your score. Relevance to the writing goal is not affected by other factors such as length.

    \item \textbf{How accurate is the feedback location (the highlighted parts it focuses on)?} \\
    \textit{Not at all $\cdot$ Slightly $\cdot$ Moderately $\cdot$ Mostly $\cdot$ Fully} \\
    We want to know if the highlight points to where the issue actually exists and isolates only the relevant text without capturing more than necessary.

    \item \textbf{How accurate is the feedback content itself with respect to the goal?} \\
    \textit{Not at all $\cdot$ Slightly $\cdot$ Moderately $\cdot$ Mostly $\cdot$ Fully} \\
    We want to know if the feedback is factually correct. Accuracy to the writing goal is not affected by other factors such as length.

    \item \textbf{How specific is this feedback in the level of detail it provides?} \\
    \textit{Poor $\cdot$ Fair $\cdot$ Good $\cdot$ Very Good $\cdot$ Excellent} \\
    We want to know if the feedback points out the exact issue and explains why it is a problem. Vague feedback that only gives a high-level issue should be rated as poor, even if it is long. Specificity is not affected by other factors such as length.

    \item \textbf{How clear is this feedback?} \\
    \textit{Poor $\cdot$ Fair $\cdot$ Good $\cdot$ Very Good $\cdot$ Excellent} \\
    We want to know if a high school student would understand this feedback without confusion, such as whether there are words that a high school student might not know. Excessively long or wordy feedback that makes the message harder to extract should be penalized.

    \item \textbf{How actionable is this feedback for revising the writing?} \\
    \textit{Poor $\cdot$ Fair $\cdot$ Good $\cdot$ Very Good $\cdot$ Excellent} \\
    We want to know if the student would know what changes to make to improve their essay after reading this. Actionable means the student writer could take a writing action in response to the feedback. Actionability is not affected by other factors such as length.

    \item \textbf{How would you describe the overall tone of this feedback?} \\
    \textit{Very negative $\cdot$ Somewhat negative $\cdot$ Neutral $\cdot$ Positive $\cdot$ Very positive} \\
    We want to know how encouraging or discouraging the feedback is. Being direct is not equivalent to being negative.
\end{enumerate}

\begin{figure*}
    \centering
    \includegraphics[width=1\linewidth]{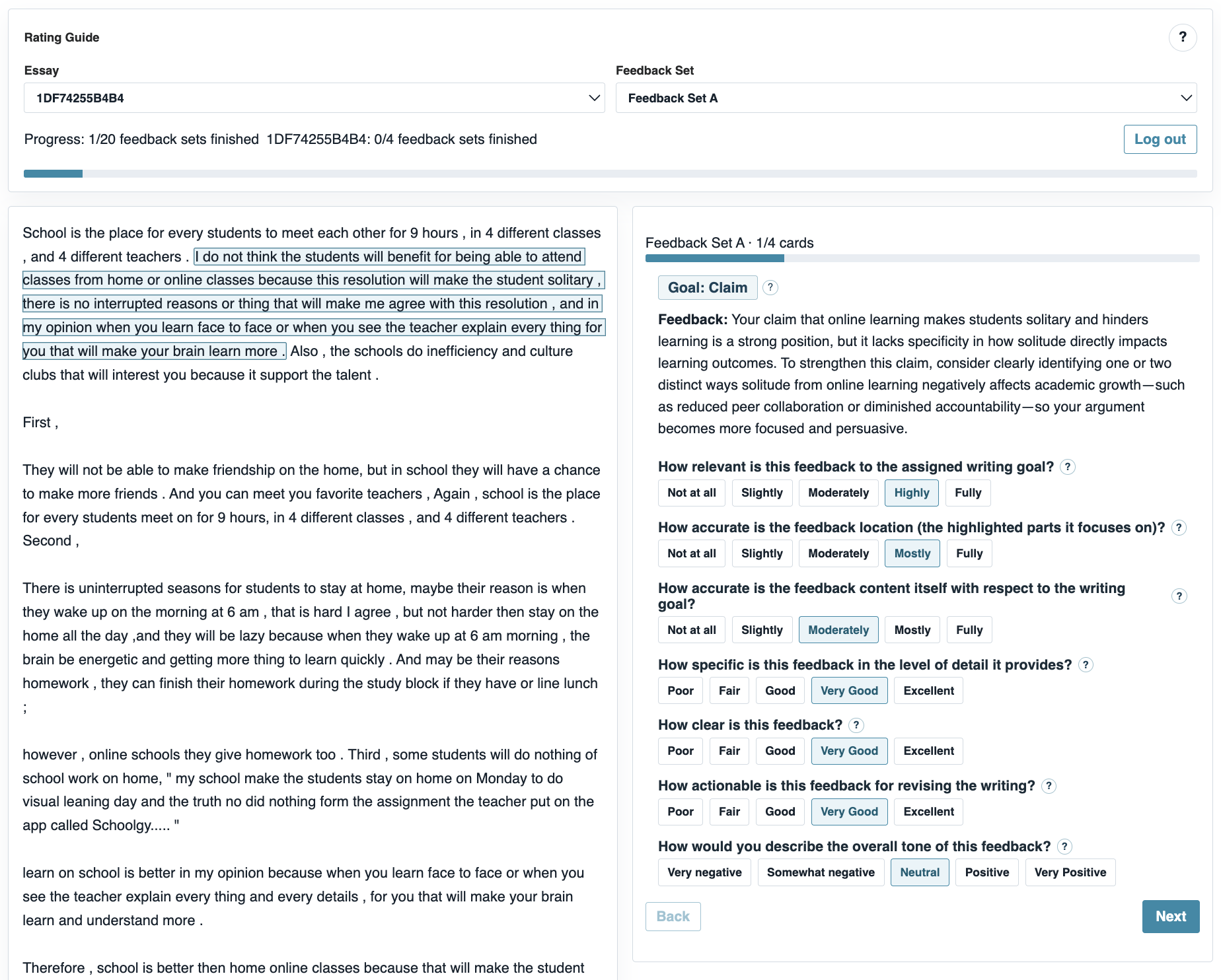}
    \caption{Rater interface for evaluating feedback quality. The left panel displays the student essay with the feedback-relevant text highlighted, while the right panel shows one feedback item, its identified writing goal, and rating questions across seven dimensions: relevance, location accuracy, content accuracy, specificity, clarity, actionability, and tone.}
    \label{fig:interface}
\end{figure*}

\subsubsection{Length-Adjusted Quality Ratings}
\label{sec:length_quality}
To separate quality differences from length differences, we fit an OLS
regression of each rubric dimension on word count for sentence-level feedback,\[D = \beta_0 + \beta_{\text{length}} \cdot \text{words} + \varepsilon,\]
on the 85 rated items (Table~\ref{tab:length_regression}). Length was a significant positive predictor of every dimension except tone (which was marginally negative, $p = .059$): one additional word in a feedback comment corresponded to an increase of $+0.033$ on specificity
(95\% CI $[0.029, 0.037]$, $R^2 = 0.75$), $+0.023$ on actionability ($R^2 = 0.54$). 

We then computed length-adjusted means by adding each side's mean residual from this model to the model's prediction at the overall mean feedback length ($\bar{\ell} \approx 64$ words). These are the predicted scores we would observe if both humans and LLMs had written feedback at the average length.

\begin{table*}[t]
\centering
\small
\begin{tabular}{lrrrrr}
\toprule
Dimension & $\beta_{\text{length}}$ & SE & $t$ & $p$ & $R^2$ \\
\midrule
Accuracy (content)  & 0.0096 & 0.0021 & 4.57  & $<.001$ & 0.20 \\
Accuracy (position) & 0.0032 & 0.0013 & 2.44  & $.017$  & 0.07 \\
Actionability       & 0.0233 & 0.0023 & 9.95  & $<.001$ & 0.54 \\
Clarity             & 0.0106 & 0.0022 & 4.76  & $<.001$ & 0.22 \\
Relevance           & 0.0062 & 0.0020 & 3.14  & $.002$  & 0.11 \\
Specificity         & 0.0331 & 0.0021 & 15.92 & $<.001$ & 0.75 \\
Tone                & $-0.0037$ & 0.0019 & $-1.92$ & $.059$ & 0.04 \\
\bottomrule
\end{tabular}
\caption{OLS regression of each rubric dimension on feedback length (in words).
For each dimension $D$, we fit $D = \beta_0 + \beta_{\text{length}} \cdot \text{words} + \varepsilon$
on the 85 rated items.
$\beta_{\text{length}}$ is the estimated change in rating per additional word;
$R^2$ is the proportion of variance in $D$ explained by length alone.}
\label{tab:length_regression}
\end{table*}

\begin{table*}[t]
\centering
\small
\begin{tabular}{lcccccc}
\toprule
& \multicolumn{3}{c}{Raw means} & \multicolumn{3}{c}{Length-adjusted means} \\
\cmidrule(lr){2-4} \cmidrule(lr){5-7}
Dimension & Human & LLM & $\Delta_{\text{raw}}$ & Human$_{\text{adj}}$ & LLM$_{\text{adj}}$ & $\Delta_{\text{adj}}$ \\
\midrule
Accuracy (content)  & 3.71 & 4.35 & $+0.65$ & 3.95 & 4.19 & $+0.24$ \\
Accuracy (position) & 4.19 & 4.38 & $+0.19$ & 4.27 & 4.33 & $+0.06$ \\
Actionability       & 2.21 & 3.61 & $+1.40$ & 2.80 & 3.21 & $+0.42$ \\
Clarity             & 2.85 & 3.42 & $+0.57$ & 3.12 & 3.24 & $+0.12$ \\
Relevance           & 3.93 & 4.35 & $+0.43$ & 4.08 & 4.25 & $+0.17$ \\
Specificity         & 1.96 & 3.80 & $+1.85$ & 2.80 & 3.24 & $+0.45$ \\
Tone                & 3.35 & 3.17 & $-0.19$ & 3.26 & 3.23 & $-0.03$ \\
\bottomrule
\end{tabular}
\caption{Raw and length-adjusted means by rubric dimension.
Length-adjusted means are computed by adding each side's mean residual from the
length-only regression (Table~\ref{tab:length_regression}) to the model's
prediction at the overall mean feedback length ($\bar{\ell} \approx 64$ words),
giving the predicted rating if both sides had written feedback at the same
average length.
$\Delta = \text{LLM} - \text{Human}$.}
\label{tab:length_adjusted}
\end{table*}

\begin{table}[t]
\centering
\small
\begin{tabular}{lcc}
\toprule
Dimension & Model $\Delta$ (SE) & $p$ (Holm) \\
\midrule
\multicolumn{3}{l}{\textit{Global-level ratings}} \\
Accuracy (content)  & $+0.70$ (0.26) & $.018^{*}$    \\
Actionability       & $+1.05$ (0.23) & $<.001^{***}$ \\
Clarity             & $+0.60$ (0.25) & $.036^{*}$    \\
Relevance           & $+0.20$ (0.14) & $.144$ (ns)   \\
Specificity         & $+1.65$ (0.35) & $<.001^{***}$ \\
Tone                & $-0.75$ (0.23) & $.005^{**}$   \\
\midrule
\multicolumn{3}{l}{\textit{Sentence-level ratings}} \\
Accuracy (position) & $+0.20$ (0.08) & $.033^{*}$    \\
Accuracy (content)  & $+0.64$ (0.11) & $<.001^{***}$ \\
Actionability       & $+1.40$ (0.12) & $<.001^{***}$ \\
Clarity             & $+0.54$ (0.12) & $<.001^{***}$ \\
Relevance           & $+0.42$ (0.11) & $<.001^{***}$ \\
Specificity         & $+1.83$ (0.13) & $<.001^{***}$ \\
Tone                & $-0.20$ (0.10) & $.056$ (ns)   \\
\bottomrule
\end{tabular}
\caption{Linear mixed-effects estimates of the LLMs $-$ Human
difference ($\Delta$) for each rating dimension, with standard
errors and Holm-corrected $p$-values ($^{*}p < .05$;
$^{**}p < .01$; $^{***}p < .001$; ns = not significant).}
\label{tab:mixed-effects}
\end{table}

\end{document}